\documentclass[journal]{IEEEtran}
\usepackage[utf8]{inputenc}
\usepackage{mathtools,lipsum,cuted}
\usepackage{amsmath,amsfonts}
\usepackage{enumitem}
\usepackage{algorithmic}
\usepackage{array}
\usepackage{subfigure}
\usepackage{stfloats}
\usepackage{url}
\usepackage{bm}
\usepackage{verbatim}
\usepackage{graphicx}
\usepackage{booktabs}
\usepackage{booktabs}
\usepackage{enumerate}
\renewcommand\eqref[1]{(\autoref{#1})}
\usepackage{mathtools,lipsum,cuted}
\usepackage{amsmath}     
\usepackage{amssymb}                       
\usepackage{mathrsfs}    
\usepackage{upgreek}
\usepackage{bm}     
\usepackage{authblk}
\usepackage{setspace}
\usepackage{graphicx}
\usepackage{amsfonts,amssymb}
\usepackage{caption}
\graphicspath{ {./figures/} }
\graphicspath{ {./Author/} }
\usepackage{amsmath}
\usepackage{cuted}
\usepackage{lineno}
\usepackage{colortbl}
\usepackage{bbding}
\usepackage{multirow}
\usepackage{makecell}
\usepackage{blindtext}
\usepackage{xcolor}
\usepackage{color}
\usepackage{setspace}
\usepackage{makecell}
\usepackage{threeparttable}
\usepackage{hyperref}
\usepackage[section]{placeins}
\usepackage{hhline}
\usepackage{threeparttable}
\usepackage{tikz}
\usepackage{diagbox}
\usepackage{pifont}
\usepackage{algorithm,algorithmic}

\hypersetup{
	colorlinks=true,
	linkcolor=cyan,
	filecolor=blue,      
	urlcolor=black,
	citecolor=green,
}
\DeclareMathAlphabet\mathbfcal{OMS}{cmsy}{b}{n}
\begin{document}

\title{Unmixing-Guided Spatial-Spectral Mamba with Clustering Tokens for Hyperspectral Image Classification}
\author{Yimin Zhu, Lincoln Linlin Xu, ~\IEEEmembership{Member,~IEEE}
\thanks{This work was supported by the Natural
Sciences and Engineering Research Council of Canada (NSERC) under Grant RGPIN-2019-06744.}
\thanks{Authors are all with the Department of Geomatics Engineering, University of Calgary, Canada (email: yimin.zhu@ucalgary.ca), Corresponding author: Lincoln Linlin Xu, lincoln.xu@ucalgary.ca}
}

\markboth{Journal of \LaTeX\ Class Files,~Vol.~13, No.~9, September~2014}%
{Shell \MakeLowercase{\textit{et al.}}: Bare Demo of IEEEtran.cls for Journals}

\maketitle

\begin{abstract}
Although hyperspectral image (HSI) classification is critical for supporting various environmental applications, it is a challenging task due to the spectral-mixture effect, the spatial-spectral heterogeneity and the difficulty to preserve class boundaries and details. This letter presents a novel unmixing-guided spatial-spectral Mamba with clustering tokens for improved HSI classification, with the following contributions. First, to disentangle the spectral mixture effect in HSI for improved pattern discovery, we design a novel spectral unmixing network that not only automatically learns endmembers and abundance maps from HSI but also accounts for endmember variabilities. Second, to generate Mamba token sequences, based on the clusters defined by abundance maps, we design an efficient Top-\textit{K} token selection strategy to adaptively sequence the tokens for improved representational capability. Third, to improve spatial-spectral feature learning and detail preservation, based on the Top-\textit{K} token sequences, we design a novel unmixing-guided spatial-spectral Mamba module that greatly improves traditional Mamba models in terms of token learning and sequencing. Fourth, to learn simultaneously the endmember-abundance patterns and classification labels, a multi-task scheme is designed for model supervision, leading to a new unmixing-classification framework that outputs not only accurate classification maps but also a comprehensive spectral-library and abundance maps. Comparative experiments on four HSI datasets demonstrate that our model can greatly outperform the other state-of-the-art approaches. Code is available at \url{https://github.com/GSIL-UCalgary/Unmixing_guided_Mamba.git}.

\end{abstract}


\IEEEpeerreviewmaketitle
\section{Introduction}

\IEEEPARstart{H}yperspectral image (HSI) classification, which transforms raw HSI data into valuable maps, is a fundamental task that supports various key environmental monitoring and resource exploration applications. Nevertheless, efficient HSI classification is challenging due to several key HSI characteristics, e.g., the spectral-mixture effect, the
spatial-spectral heterogeneity, and the difficulty in preserving class
boundaries and details. Given these difficulties, it is critical to extract discriminative features that can efficiently capture subtle differences among HSI classes. 

Spectral unmixing aims to disentangle the mixed pixels into pure spectral signal (i.e., endmembers) and their fractional abundanaces, which is critical for addressing the HSI spatial-spectral heterogeneity issue and discovering subtle and valuable patterns in HSI. Therefore, integrating spectral unmixing with HSI classification might significantly improve the classification performance. Although some related works \cite{10856229,10570241} fused the abundance and classification features at the final decision level, there are still research gaps in integrating spectral unmixing with HSI classification within a more coherent framework via advanced deep learning approaches. 


\begin{figure}[!t]
    \centering
    \includegraphics[width=0.5\textwidth]{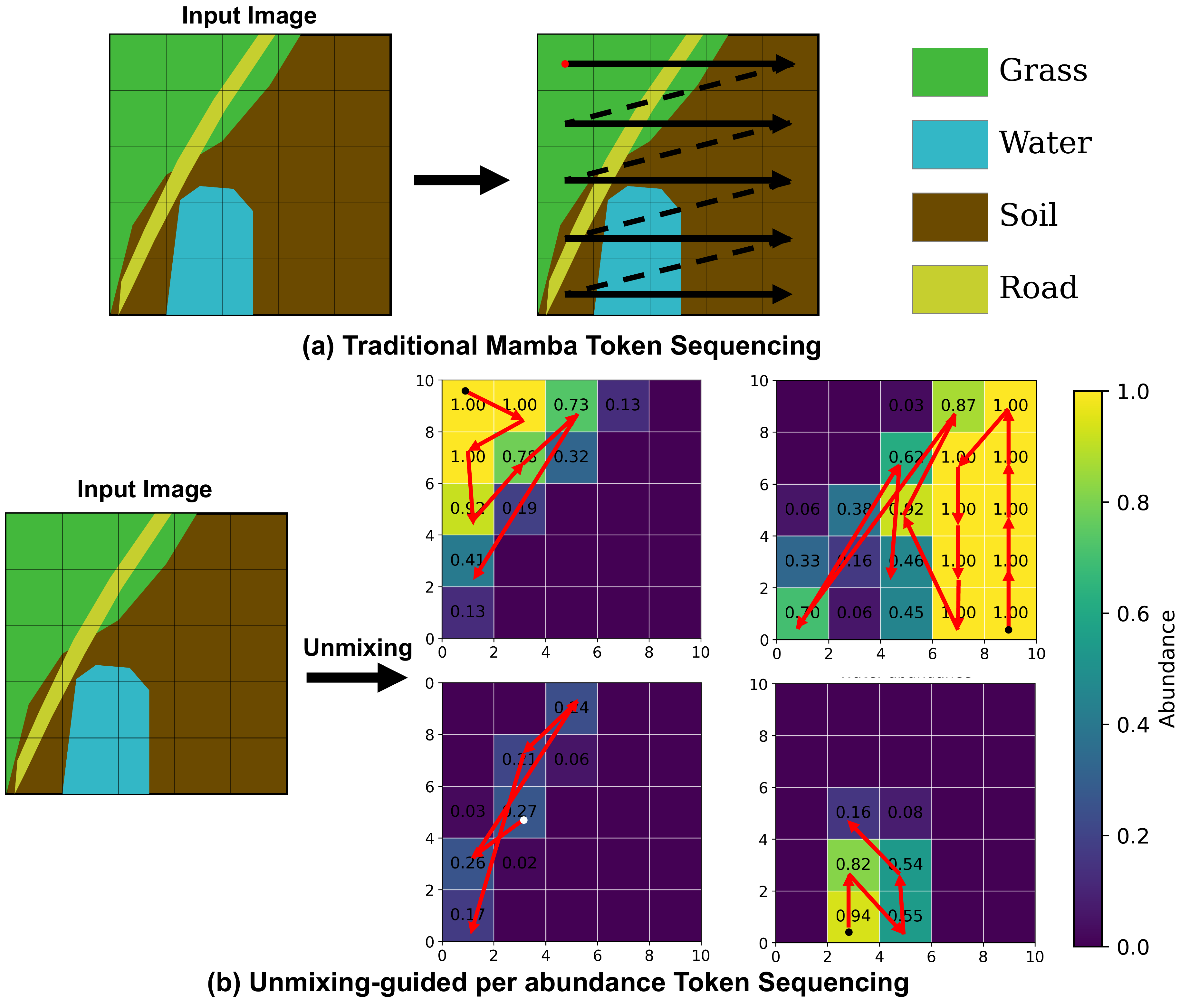}
    \caption{\textbf{Traditional Mamba token sequencing} treats an entire image or uses a pre-defined way—despite its highly heterogeneous spatial patterns—as a single long sequence, which suffers from being inefficient at modeling fine-scale structures. In contrast, our \textbf{Unmixing-guided per-abundance token sequencing} decomposes the hyperspectral image into physically meaningful abundance maps, which better capture subtle and weak patterns and edges, and constructs a dedicated token sequence from each abundance, leading to  \textbf{non-local, sparse, dynamic, and adaptive} Mamba token sequence.}
    \label{unmixing_guided scanning}
\end{figure}


Deep learning approaches have been widely used for HSI classification. Although Convolutional Neural Networks (CNNs) can effectively capture local spatial-spectral features \cite{SSRN}, they have limitations in terms of modeling long-range spatial-spectral dependencies. Transformers, on the other hand, although they can better address the large-scale long-range correlation effect in HSI, have a very high computational cost and risk of overfitting due to the numerous parameters in the attention mechanism \cite{sstn}. Recently, Mamba models have overcome the limitations of Transformer models by modeling the long-range and large-scale effect \cite{liu2025vision} with linear computational complexity, which not only decreases the computational cost but also increases the generalization capability. 

However, one significant limitation of the Mamba model is the notorious correlation decay issue, which is especially serious when modeling the entire HSI image using one long token sequence. Due to the heavy spatial-spectral heterogeneity and complexity in HSI images, using a single Mamba token sequence will not only increase the computational cost to process a really long sequence, but also decrease the model efficiency to capture the subtle and weak patterns in the HSI image.

Therefore, how to use spectral unmixing to break down the HSI image into less complex and physically-meaningful clusters, and then build cluster-dedicated Mamba sequences to improve the Mamba modeling efficiency is a critical research issue. 


This letter, therefore, presents a novel unmixing-guided spatial-spectral Mamba with clustering Top-\textit{K} tokens with the following contributions:

\begin{itemize}

\item To disentangle the spectral mixture effect in HSI for improved pattern discovery, we design a novel spectral unmixing network that not only automatically learns endmembers and abundance maps from HSI but also accounts for endmember variabilities. 

\item To generate Mamba token sequences, based on the clusters defined by abundance maps, we design an efficient Top-\textit{K} token selection strategy to adaptively sequence the tokens for improved representational capability. 

\item To improve spatial-spectral feature learning and detail preservation, based on the Top-\textit{K} token sequences, we design a novel unmixing-guided spatial-spectral Mamba module that greatly improves traditional Mamba models in terms of token learning and sequencing. 

\item To learn simultaneously the endmember-abundance patterns and classification labels, a multi-task scheme is designed for model supervision, leading to a new unmixing-classification framework that outputs not only accurate classification maps but also a comprehensive spectral-library and abundance maps.
\end{itemize}




The remainder of the article is organized as follows. \autoref{method_part} describes our proposed method in detail. \autoref{Experiments} shows the experimental results. \autoref{conclusion} draws the conclusions.

\section{Methodology} \label{method_part}

\subsection{Model Overview}

\begin{figure*}[!h]
    \centering
    \includegraphics[width=1.03\textwidth]{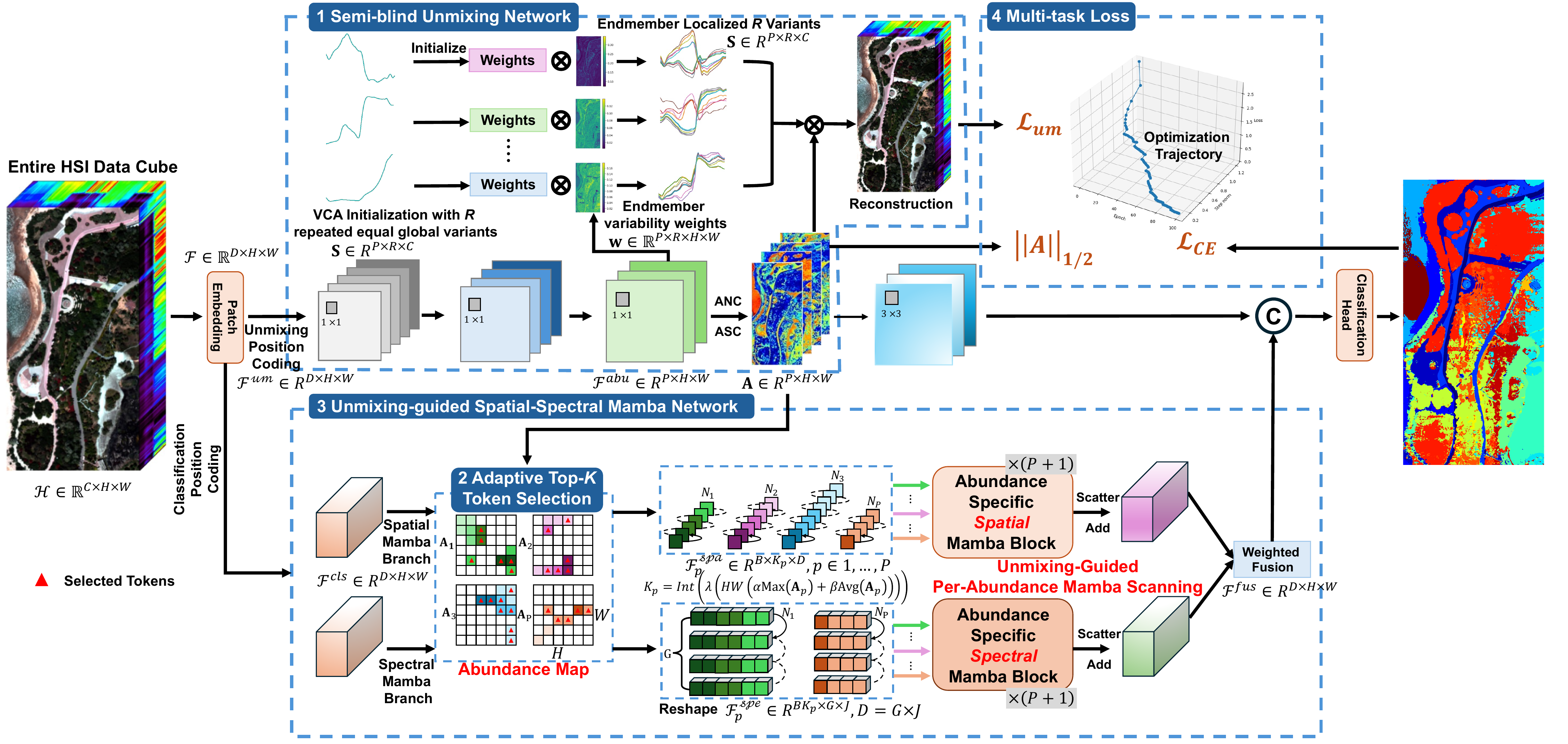}
    \caption{The proposed model contains (1) a semi-blind spectral unmixing branch for abundance learning and endmember variability modeling \autoref{Unmixing Branch}, (2) adaptive Top-\textit{K} tokens selection from abundance \autoref{tokens}, (3) the unmixing-guided spatial-spectral Mamba block for learning selected sparse tokens \autoref{Unmixing-guided}, and (4) a multi-task learning part for balancing different tasks \autoref{Multi-task}. Model outputs classification map, abundance map, and endmember with variability.}
    \label{unmixing_guided arch}
\end{figure*}

\autoref{unmixing_guided arch} shows the overall architecture of our proposed model. The whole hyperspectral image \(\mathcal{H} \in \mathbb{R}^{C \times H \times W}\) is first processed by a patch embedding layer, forming feature \(\mathcal{F} \in \mathbb{R}^{D \times H \times W}\). Given the different emphases of the classification and unmixing tasks, two different attention-like position coding mechanisms are used for each branch, yielding unmixing feature \(\mathcal{F}^{\text{um}}\in \mathbb{R}^{D \times H \times W}\) and classification feature \(\mathcal{F}^{\text{cls}}\in \mathbb{R}^{D \times H \times W}\), respectively. Two branches are connected by the abundance clusters \(\mathbf{A} \in \mathbb{R}^{P \times H \times W}\), which are used to select sparse tokens for sequencing modeling on spatial and spectral domains separately by using an adaptive token selection strategy. Features from each branch are fused to get the final prediction. A multi-task objective function is used to update the model weights. The whole model outputs a classification map, abundance, and endmember with variability. \(C\), \(H\), \(W\), \(D\), and \(P\) denote the spectral channel, spatial size of imagery, feature dimension, and number of endmembers. The unmixing module, token selection, unmixing-guided spatial–spectral Mamba module, and multi-task are illustrated in \autoref{Unmixing Branch}, \autoref{tokens}, \autoref{Unmixing-guided}, and \autoref{Multi-task} respectively.

\subsection{Unmixing Branch with Endmember Variability Learning} \label{Unmixing Branch}

\subsubsection{Abundance Extraction}
Spectral unmixing usually refers to the linear mixing model (LMM), which decompose the original hyperspectral image cube \(\mathcal{H}\) into abundance \(\mathbf{A}\) and the corresponding endmember matrix \(\mathbf{S} \in \mathbb{R}^{P \times C}\). The LMM can be expressed as:
\begin{align}
    \begin{aligned}
        \mathcal{H} = \mathbf{S}^T \mathbf{A} + \mathbf{N}, s.t. \mathbf{A} \geq 0, \mathbf{1}^{T}_{p} \mathbf{A}=\mathbf{J}
    \end{aligned}
\end{align}
where \(\mathbf{N} \in \mathbb{R}^{C \times H \times W}\) denotes the addative noise and \(\mathbf{J}\) is a matrix consisting entirely of ones. The input feature \(\mathcal{F}^{\text{um}}\) for unmixing branch is expressed as:
\begin{align}
    \begin{aligned}
    &\text{Pos}_{\text{um}} =\left[ x_{i,j}, y_{i,j}, \frac{\sin (2 \pi x_{i,j}) + \cos (2\pi y_{i,j})}{2}\right] \\
    \mathcal{Q}_{\text{um}} &= \text{Proj}_{D+3 \rightarrow D} ([\mathcal{F};\text{Pos}_{\text{um}}]), \mathcal{K}_{\text{um}} = \text{Proj}_{H \rightarrow 1}(\mathcal{F}) \\
    \mathcal{V}_{\text{um}} &= \text{Proj}_{W \rightarrow1}(\mathcal{F}), \mathcal{F}^{\text{um}} = \text{tanh}(\mathcal{Q}_{\text{um}} \odot \mathcal{K}_{\text{um}}) \odot \mathcal{V}_{\text{um}} + \mathcal{F}\\
    \end{aligned}
\end{align}
where \(x_{i,j}, y_{i,j}\) is the normalized
spatial coordinate. The abundance estimation network is designed using a \(1 \times 1\) convolution operation, which further ensures the integrity of the spatial structure of the extracted abundance maps and fully explores the spectral relationships among different ground objects. The feature obtained can be expressed as:
\begin{align}
    \begin{aligned}
        \mathcal{F}^{\text{abu}} = \text{Conv}_{1\times1}(f(\text{BN}(\text{Conv}_{1\times1}(f(\text{BN}(\text{Conv}_{1\times1}(\mathcal{F^{\text{um}}})))))))
    \end{aligned}
\end{align}
where \(f(\cdot)\) is the ReLU activation and BN is the batch normalization. Then, the estimated abundance is:
\begin{align}
    \begin{aligned}
        \mathbf{A} = {|\mathcal{F}^{\text{abu}}(p)|}/{\sum_{p=1}^P |\mathcal{F}^{\text{abu}}(p)|}, p=1,...,P
    \end{aligned}
    \label{abu_equation}
\end{align}

\subsubsection{Endmember Module with Variability Modeling}
Instead of total blind unmixing, we use vertex component analysis (VCA) to initialize the weights of the endmember module. Considering the spectral variability, we repeat each endmember derived from VCA into a set of \(R\) equal global variants \(\mathbf{S} \in \mathbb{R}^{P \times R \times C}\) and further introduce pixel-wise endmember variability weights which are learn from feature \(\mathcal{F}^{\text{abu}}\) to learn more localized variants, as shown in \autoref{unmixing_guided arch} (top). The observed spectrum at each pixel is modeled as:
\begin{align}
    \begin{aligned}
        \hat{\mathcal{H}}(h,w) = \sum_{p=1}^P a_{p}(h,w)\sum_{r=1}^R\mathbf{w}_{p,r}(h,w)\mathbf{S}_{p,r}
    \end{aligned}
    \label{endmember_variability}
\end{align}
where \(a_{p}(h,w)\) denotes the abundance of the \(p\)-th endmember and \(\mathbf{w}_{p,r}\) represents the contribution of its \(r\)-th variant. The variability weights are constrained to form a probability distribution, i.e., \(\sum_{r} \mathbf{w}_{p,r}(h,w) = 1\). This formulation can be interpreted as a pixel-wise adaptive endmember model,  where each endmember is dynamically reconstructed as a convex combination of multiple spectral variants.

\subsection{Adaptive Top-\textit{K} Token Selection from Abundance Clusters} \label{tokens}
After getting the abundance map, informative tokens need to be selected for the Mamba model processing. Inspired by \cite{han2019learning}, we choose tokens based on temporal abundance:
\begin{align}
    \begin{aligned}
        &\mathbf{A}_{\text{temp}}^t = (1-\gamma_t) \mathbf{A}^t + \gamma_t \tilde{\mathbf{A}}^{t-1}\\
        &\tilde{\mathbf{A}}^t = \tau \tilde{\mathbf{A}}^{t-1} + (1-\tau) \mathbf{A}^t
    \end{aligned}
\end{align}
where \(t\) is the epoch, \(\mathbf{A}^t\) is the output of \autoref{abu_equation} at epoch \(t\), \(\tilde{\mathbf{A}}^t\) is the temporal abundance using EMA updating. \(\tau\) is the momentum coefficient. \(\gamma_t\) is the blending weight. Temporal abundance embedding provides a history-aware regularization that stabilizes abundance estimation and ensures consistent token selection.
Based on this, we then use a simple but efficient way to determine the number of tokens:
\begin{align}
    \begin{aligned}
        &K_{p} = \text{Int}(\lambda (H W (\alpha \text{Max}(\mathbf{A}_{\text{temp}}^t(p)) + \beta \text{Avg}(\mathbf{A}_{\text{temp}}^t(p))))
    \end{aligned}
    \label{Np}
\end{align}
where \(\lambda\) controls the length of the token, which we set 0.1 for each dataset. \(\text{Max}(\mathbf{A}_{\text{temp}}^t(p))\) focuses on the peak value for \(p\)-th abundance clusters, while \(\text{Avg}(\mathbf{A}_{\text{temp}}^t(p))\) focuses more on the spatial diversity, high value means that the abundance be more distributed, considering the sparisty of abundance. We set \(\alpha=0.3\) and \(\beta=0.7\). This makes \(\sum_{p=1}^P K_p \ll HW\).  For example, for a \(512 \times 512 \times 200\) hyperspectral image, yielding \(L=HW=512^2=262144\) tokens, then the complexity is \(\mathcal{O}(Lds)\), where \(d\) and \(s\) demote the hidden dimension and state size, respectively. In contrast, the proposed model only processes informative tokens from each abundance map, and then the complexity is \(\mathcal{O}((\sum_{p=1}^PK_p)ds\)), where \(\sum_{p=1}^P K_P \ll L\), since sparse tokens \(N_p\) are selected from \(p\)-th abundance map.

\subsection{Unmixing-Guided Spatial-Spectral Mamba with
Clustering Tokens} \label{Unmixing-guided}






In the Spatial-Spectral Mamba Block, a series of parallel small mamba layers is used to learn the relationship among selected tokens. This way makes the model wider. Total \(P+1\) parallel mamba blocks in the Spatial and Spectral Mamba, the additional one is used for those unselected pixels, which are also selected sparsely. The input feature of the classification branch \(\mathcal{F}^{\text{cls}}\) is expressed as:
\begin{align}
    \begin{aligned}
    &\text{Pos}_{\text{cls}} =\left[ \sin (2 \pi x_{i,j}), \cos (2\pi y_{i,j}), \frac{x_{i,j}+y_{i,j}}{2}\right] \\
    \mathcal{Q}_{\text{cls}} &= \text{Proj}_{D+3 \rightarrow D} ([\mathcal{F};\text{Pos}_{\text{cls}}]), \mathcal{K}_{\text{cls}} = \text{Proj}_{D \rightarrow 1}(\mathcal{F}) \\
    \mathcal{V}_{\text{cls}} & = \text{Proj}_{D \rightarrow1}(\mathcal{F}), \mathcal{F}^{\text{cls}} = \text{tanh}(\mathcal{Q}_{\text{cls}} \odot \mathcal{K}_{\text{cls}}) \odot \mathcal{V}_{\text{cls}} + \mathcal{F}\\
    \end{aligned}
    \label{F_cls}
\end{align}

\paragraph{Abundance-specific Spatial Mamba Block} 
Given the input feature \(\mathcal{F}^{\text{cls}}\) and the temporal abundance map \(\mathbf{A}_{\text{temp}}^t(p)\), abundance-specific spatial tokens are adaptively selected from \(\mathcal{F}^{\text{cls}}\). The selected spatial tokens \(\mathcal{F}_p^{\text{spa}} \in \mathbb{R}^{B \times K_p \times D}\) on \(p\)-th abundance map are expressed as follows:
\begin{align}
    \begin{aligned}
        \mathcal{F}_p^{\text{spa}} = \mathcal{S}(\Omega(\mathcal{F}^{\text{cls}}, \mathbf{A}_{\text{temp}}^t(p), K_p))
    \end{aligned}
    \label{spa_tokens}
\end{align}
\(\Omega\) is the function that picks the \(N_p\) tokens, and \(\mathcal{S}\) is the function to sort the tokens from high abundance value to low value. Since there is a total \(P\) abundance map, meaning that \(P\) spatial mamba blocks are built to learning the abundance-specific spatial information \(\mathcal{F}_p^{\text{spa}}\) in parallel, as well as the tokens that haven't been selected. The final results are scattered back to the original location and residual connected with the input feature \(\mathcal{F}^{\text{cls}}\). \(B\) is the batch size, here it is 1.

\paragraph{Abundance-specific Spectral Mamba Block}
In the Abundance-specific Spatial Mamba Block, the spatial token definition is \(\mathcal{F}_p^{\text{spa}} \in \mathbb{R}^{B \times K_p \times D}\), while in the Abundance-specific Spectral Mamba Block, the token will be reshaped to \(\mathcal{F}_p^{\text{spe}} \in \mathbb{R}^{BK_p \times G \times J}, D=G \times J, G=4\). This allows the model to learn the grouped spectral domain features across all the selected informative tokens. Similar to \autoref{spa_tokens}, \(\mathcal{F}_p^{\text{spe}}\) is formulated as follows:
\begin{align}
    \begin{aligned}
        \mathcal{F}_p^{\text{spe}} = \text{Reshape}(\mathcal{S}(\Omega(\mathcal{F}^{\text{cls}}, \mathbf{A}_{\text{temp}}^t(p), N_p)))
    \end{aligned}
\end{align}
Also, \(P\) number of Abundance-specific Spectral Mamba Blocks are performed in parallel, and another Mamba block is used for the unselected tokens. Finally, the fused feature \(\mathcal{F}^{\text{fus}}\) from the spectral and spatial branches is concatenated with the abundance map feature, and then fed into the classification head to get the prediction.

\subsection{Multi-task Loss Function} \label{Multi-task}
The final objective function is combined with the unmixing and the classification task, forming a multi-task loss function:
\begin{align}
    \begin{aligned}
        \mathcal{L}_{\text{total}} = \mathcal{L}_{\text{CE}} + 0.01  \mathcal{L}_{\text{um}} + 0.001 ||\mathbf{A}||_{1/2}
    \end{aligned}
    \label{multi-loss}
\end{align}
where \(\mathcal{L}_{\text{CE}}\) is the cross-entropy loss. \(\mathcal{L}_{\text{um}}\) is the reconstruction for the unmixing task using spectral angle distance (SAD), \(\mathcal{L}_{\text{SAD}}=\frac{1}{HW}\sum_{h=1}^H \sum_{w=1}^W \arccos \left[{\frac{\mathcal{H}_{h,w}^T \hat{\mathcal{H}}_{h,w}^T}{||\mathcal{H}_{h,w}||_2 ||\hat{\mathcal{H}}_{h,w}||_2}}\right]\). \(||\mathbf{A}||_{1/2}\) is non-convex sparse penalty on abundance.

\section{Experiments and Analysis} \label{Experiments}

\subsection{Implement Schema}
We compare with various approaches, including Random Forest, SSRN \cite{SSRN}, SS-ConvNeXt \cite{ssconvnext}, SSTN \cite{sstn}, DSNet \cite{10570241}, S\(^{2}\)VNet \cite{10856229}, MambaHSI \cite{10604894}, SFMamba \cite{meng2026spatial}, SDMamba \cite{11075710} on four datasets, i.e., QUH-Tangdaowan\footnote{\label{fn:tangdaowan}\url{https://github.com/RsAI-lab/QUH-classification-dataset}}, Houston, Pavia University, and Fanglu Tea Farm \footnote{\label{Fanglu}\url{https://www.geodoi.ac.cn/weben/doi.aspx?Id=720}}. Note that DSNet, S\(^{2}\)VNet are methods with unmixing assistance, and S\(^{2}\)VNet considers the spectral variability. All the datasets use 1\% per class as training samples, 1\% for validation, and the rest for testing. The learning rate is empirically set to \(1e\)-3 and \(5e\)-4 for classification and unmixing, respectively, and decays by a factor of 0.9 every 50 epochs, with a total of 500 epochs. \(P\) is set to 23, 14, 25, and 15 for the four datasets, respectively, see \autoref{endmember_P}. Our model uses the entire image as input. 
\begin{table}[]
\caption{Performance on the four datasets.}
\resizebox{0.49\textwidth}{!}{
\begin{tabular}{cccccccccccccccc}
\hline \hline 
\multicolumn{16}{c}{QUH-Tangdaowan}                                                                                                                                                                                                                                                                                                                                                                                        \\ \hline
\multirow{2}{*}{Class No.} & \multicolumn{1}{c|}{\multirow{2}{*}{Class Name}} & \multicolumn{1}{c|}{\multirow{2}{*}{Train Num}} & \multicolumn{9}{c|}{Methods}                                                                                                                                                & \multicolumn{4}{c}{\textbf{Ablation Study}}                                                                         \\ \cline{4-16} 
                           & \multicolumn{1}{c|}{}                            & \multicolumn{1}{c|}{}                           & RF     & SSRN            & SS-ConvNeXt     & DSNet           & S2VNet          & MambaHSI        & SFMamba         & SDMamba         & \multicolumn{1}{c|}{Ours}            & \multicolumn{1}{l}{W/O \(\text{Pos}_{\text{um}}\)} & \multicolumn{1}{l}{W/O \(\text{Pos}_{\text{cls}}\)} & W/O Top-\textit{K} & \multicolumn{1}{l}{W/O Variability} \\ \hline
1                          & \multicolumn{1}{c|}{Rubber track}                & \multicolumn{1}{c|}{258}                        & 97.70  & 99.38           & 99.29           & 99.27           & 99.85           & \cellcolor[RGB]{254, 248, 198}\textbf{99.94}  & 99.71           & 99.64           & \multicolumn{1}{c|}{99.72}           & 99.72                      & 99.55                       & 99.56     & 99.81                               \\
2                          & \multicolumn{1}{c|}{Flagging}                    & \multicolumn{1}{c|}{555}                        & 92.94  & 97.02           & 97.85           & 97.54           & 97.72           & 99.59           & 98.78           & 97.62           & \multicolumn{1}{c|}{\cellcolor[RGB]{254, 248, 198}\textbf{99.18}}  & 98.45                      & 98.60                       & 77.63     & 99.14                               \\
3                          & \multicolumn{1}{c|}{Sandy}                       & \multicolumn{1}{c|}{340}                        & 87.41  & 94.04           & 93.60           & 90.71           & 94.33           & 98.09           & 95.81           & 94.39           & \multicolumn{1}{c|}{\cellcolor[RGB]{254, 248, 198}\textbf{98.21}}  & 98.23                      & 96.06                       & 93.03     & 98.04                               \\
4                          & \multicolumn{1}{c|}{Asphalt}                     & \multicolumn{1}{c|}{606}                        & 96.03  & 99.05           & 99.42           & 99.20           & 99.40           & \cellcolor[RGB]{254, 248, 198}\textbf{99.80}  & 99.33           & 99.35           & \multicolumn{1}{c|}{99.18}           & 99.49                      & 98.93                       & 91.96     & 99.49                               \\
5                          & \multicolumn{1}{c|}{Boardwalk}                   & \multicolumn{1}{c|}{18}                         & 50.60  & 91.83           & 91.39           & 90.57           & 91.88           & 96.22           & \cellcolor[RGB]{254, 248, 198}\textbf{96.87}  & 67.71           & \multicolumn{1}{c|}{94.96}           & 91.07                      & 79.03                       & 0.00      & 96.00                               \\
6                          & \multicolumn{1}{c|}{Rocky shallows}              & \multicolumn{1}{c|}{371}                        & 80.64  & 86.86           & 85.93           & 86.41           & 87.88           & \cellcolor[RGB]{254, 248, 198}\textbf{99.28}  & 87.57           & 88.16           & \multicolumn{1}{c|}{99.21}           & 98.89                      & 96.27                       & 95.14     & 99.02                               \\
7                          & \multicolumn{1}{c|}{Grassland}                   & \multicolumn{1}{c|}{141}                        & 50.16  & 77.30           & 74.43           & 69.02           & 75.76           & 89.18           & \cellcolor[RGB]{254, 248, 198}\textbf{89.58}  & 72.73           & \multicolumn{1}{c|}{89.83}           & 88.39                      & 80.59                       & 73.95     & 87.75                               \\
8                          & \multicolumn{1}{c|}{Bulrush}                     & \multicolumn{1}{c|}{640}                        & 98.48  & 99.73           & 99.77           & 99.63           & 99.94           & 99.89           & \cellcolor[RGB]{254, 248, 198}\textbf{100.00} & 99.54           & \multicolumn{1}{c|}{99.89}           & 99.77                      & 99.57                       & 98.85     & 99.90                               \\
9                          & \multicolumn{1}{c|}{Gravel road}                 & \multicolumn{1}{c|}{306}                        & 89.43  & 97.95           & 97.19           & 98.21           & 98.53           & 98.61           & \cellcolor[RGB]{254, 248, 198}\textbf{99.77}  & 97.22           & \multicolumn{1}{c|}{97.28}           & 98.28                      & 97.16                       & 39.87     & 97.51                               \\
10                         & \multicolumn{1}{c|}{Ligustrum vicaryi}           & \multicolumn{1}{c|}{17}                         & 70.92  & 82.71           & 91.92           & 89.86           & 91.81           & 98.97           & 99.42           & 91.13           & \multicolumn{1}{c|}{\cellcolor[RGB]{254, 248, 198}\textbf{99.43}}  & 98.97                      & 92.51                       & 84.90     & 96.91                               \\
11                         & \multicolumn{1}{c|}{Coniferous pine}             & \multicolumn{1}{c|}{212}                        & 33.96  & 55.75           & 69.14           & 59.98           & 57.33           & 92.00           & 82.64           & 60.11           & \multicolumn{1}{c|}{\cellcolor[RGB]{254, 248, 198}\textbf{98.93}}  & 98.77                      & 94.46                       & 87.64     & 98.92                               \\
12                         & \multicolumn{1}{c|}{Spiraea}                     & \multicolumn{1}{c|}{7}                          & 9.11   & 67.34           & 3.80            & 46.12           & 45.98           & 92.24           & 91.70           & 44.75           & \multicolumn{1}{c|}{\cellcolor[RGB]{254, 248, 198}\textbf{94.29}}  & 90.48                      & 57.14                       & 53.06     & 96.33                               \\
13                         & \multicolumn{1}{c|}{Bare soil}                   & \multicolumn{1}{c|}{16}                         & 81.84  & 99.15           & 99.63           & 95.64           & 98.24           & 99.57           & 98.48           & \cellcolor[RGB]{254, 248, 198}\textbf{100.00} & \multicolumn{1}{c|}{99.88}           & 100.00                     & 96.01                       & 87.90     & 100.00                              \\
14                         & \multicolumn{1}{c|}{Buxus sinica}                & \multicolumn{1}{c|}{8}                          & 65.09  & 86.63           & 74.42           & 73.15           & 71.42           & 65.97           & \cellcolor[RGB]{254, 248, 198}\textbf{91.35}  & 82.14           & \multicolumn{1}{c|}{86.21}           & 84.60                      & 71.84                       & 28.73     & 91.84                               \\
15                         & \multicolumn{1}{c|}{Photinia serrulata}          & \multicolumn{1}{c|}{140}                        & 73.63  & 81.71           & 81.04           & 80.80           & 77.42           & \cellcolor[RGB]{254, 248, 198}\textbf{90.13}  & 88.02           & 84.91           & \multicolumn{1}{c|}{88.09}           & 85.47                      & 83.54                       & 72.96     & 86.43                               \\
16                         & \multicolumn{1}{c|}{Populus}                     & \multicolumn{1}{c|}{1409}                       & 94.22  & 94.54           & 94.84           & 92.34           & 95.14           & 96.56           & 97.02           & 94.70           & \multicolumn{1}{c|}{\cellcolor[RGB]{254, 248, 198}\textbf{97.58}}  & 97.87                      & 95.50                       & 96.06     & 97.97                               \\
17                         & \multicolumn{1}{c|}{Ulmus pumila L}              & \multicolumn{1}{c|}{98}                         & 52.90  & 82.58           & 85.68           & 78.44           & 82.56           & 91.96           & 91.51           & 82.15           & \multicolumn{1}{c|}{\cellcolor[RGB]{254, 248, 198}\textbf{96.25}}  & 98.40                      & 90.18                       & 53.88     & 95.99                               \\
18                         & \multicolumn{1}{c|}{Seawater}                    & \multicolumn{1}{c|}{422}                        & 99.19  & 99.55           & \cellcolor[RGB]{254, 248, 198}\textbf{100.00} & 99.98           & 99.99           & 99.96           & 99.19           & \textbf{100.00} & \multicolumn{1}{c|}{99.76}           & 99.84                      & 99.67                       & 93.90     & 99.89                               \\ \hline
\multicolumn{3}{c|}{OA}                                                                                                         & 88.66  & 93.59           & 94.10           & 92.72           & 93.94           & 97.76           & 96.46           & 93.86           & \multicolumn{1}{c|}{\cellcolor[RGB]{254, 248, 198}\textbf{98.25}}  & 98.11                      & 96.36                       & 88.01     & 98.19                               \\
\multicolumn{3}{c|}{AA}                                                                                                         & 73.57  & 88.51           & 85.52           & 85.94           & 86.96           & 94.89           & 94.82           & 86.46           & \multicolumn{1}{c|}{\cellcolor[RGB]{254, 248, 198}\textbf{96.54}}  & 95.92                      & 90.36                       & 73.84     & 96.71                               \\
\multicolumn{3}{c|}{Kappa(\%)}                                                                                                  & 86.96  & 92.74           & 93.28           & 91.70           & 93.08           & 97.21           & 96.00           & 92.98           & \multicolumn{1}{c|}{\cellcolor[RGB]{254, 248, 198}\textbf{98.01}}  & 97.85                      & 95.86                       & 86.28     & 97.93                               \\ \hline \hline 
\multicolumn{16}{c}{Pavia University}                                                                                                                                                                                                                                                                                                                                                                                      \\ \hline
1                          & \multicolumn{1}{c|}{Asphalt}                     & \multicolumn{1}{c|}{66}                         & 92.84  & 96.30           & 97.33           & 98.33           & 97.64           & 99.01           & 89.98           & \cellcolor[RGB]{254, 248, 198}\textbf{98.65}  & \multicolumn{1}{c|}{98.64}           & 98.45                      & 91.95                       & 99.14     & 98.17                               \\
2                          & \multicolumn{1}{c|}{Meadows}                     & \multicolumn{1}{c|}{186}                        & 98.33  & 98.49           & 97.32           & \cellcolor[RGB]{254, 248, 198}\textbf{99.68}  & 99.00           & 99.66           & 97.70           & 99.56           & \multicolumn{1}{c|}{99.20}           & 98.81                      & 98.65                       & 98.98     & 98.92                               \\
3                          & \multicolumn{1}{c|}{Gravel}                      & \multicolumn{1}{c|}{20}                         & 27.51  & 90.86           & 93.84           & 70.63           & 89.30           & 83.43           & 95.91           & 82.89           & \multicolumn{1}{c|}{\cellcolor[RGB]{254, 248, 198}\textbf{95.63}}  & 97.33                      & 96.65                       & 86.69     & 95.39                               \\
4                          & \multicolumn{1}{c|}{Trees}                       & \multicolumn{1}{c|}{30}                         & 81.54  & 96.76           & 95.36           & 92.90           & 96.36           & 84.65           & \cellcolor[RGB]{254, 248, 198}\textbf{97.20}  & 95.20           & \multicolumn{1}{c|}{96.17}           & 95.91                      & 96.87                       & 95.47     & 95.91                               \\
5                          & \multicolumn{1}{c|}{Painted metal sheets}        & \multicolumn{1}{c|}{13}                         & 97.57  & 97.62           & 97.77           & 99.84           & 93.54           & \cellcolor[RGB]{254, 248, 198}\textbf{100.00} & 99.62           & 99.70           & \multicolumn{1}{c|}{98.94}           & 98.94                      & 96.21                       & 99.09     & 98.94                               \\
6                          & \multicolumn{1}{c|}{Bare Soil}                   & \multicolumn{1}{c|}{50}                         & 36.29  & 97.67           & 95.95           & 98.62           & 99.53           & 99.97           & 99.93           & 93.55           & \multicolumn{1}{c|}{\cellcolor[RGB]{254, 248, 198}\textbf{100.00}} & 99.78                      & 89.13                       & 99.94     & 99.57                               \\
7                          & \multicolumn{1}{c|}{Bitumen}                     & \multicolumn{1}{c|}{13}                         & 59.50  & 94.55           & 94.23           & 87.80           & 91.71           & 87.11           & \cellcolor[RGB]{254, 248, 198}\textbf{100.00} & 80.14           & \multicolumn{1}{c|}{95.71}           & 95.71                      & 94.25                       & 91.64     & 93.71                               \\
8                          & \multicolumn{1}{c|}{Self-Blocking Bricks}        & \multicolumn{1}{c|}{36}                         & 86.91  & 94.00           & 91.81           & 88.55           & 94.00           & 95.84           & 96.34           & 93.74           & \multicolumn{1}{c|}{\cellcolor[RGB]{254, 248, 198}\textbf{98.06}}  & 96.86                      & 92.66                       & 97.51     & 96.32                               \\
9                          & \multicolumn{1}{c|}{Shadows}                     & \multicolumn{1}{c|}{9}                          & 100.00 & 97.42           & 96.37           & \cellcolor[RGB]{254, 248, 198}\textbf{100.00} & 97.70           & 94.51           & 96.87           & 98.49           & \multicolumn{1}{c|}{98.92}           & 99.14                      & 99.68                       & 99.46     & 97.09                               \\ \hline
\multicolumn{1}{l}{}       & \multicolumn{2}{c|}{OA}                                                                            & 83.33  & 96.43           & 96.14           & 96.13           & 96.36           & 96.90           & 96.62           & 96.46           & \multicolumn{1}{c|}{\cellcolor[RGB]{254, 248, 198}\textbf{98.59}}  & 98.33                      & 95.55                       & 97.92     & 98.06                               \\
\multicolumn{1}{l}{}       & \multicolumn{2}{c|}{AA}                                                                            & 75.61  & 95.96           & 95.55           & 92.93           & 95.42           & 93.80           & 97.06           & 93.54           & \multicolumn{1}{c|}{\cellcolor[RGB]{254, 248, 198}\textbf{97.92}}  & 97.88                      & 95.11                       & 96.43     & 97.11                               \\
\multicolumn{1}{l}{}       & \multicolumn{2}{c|}{Kappa(\%)}                                                                     & 77.02  & 95.54           & 95.95           & 94.85           & 96.32           & 95.58           & 95.56           & 95.28           & \multicolumn{1}{c|}{\cellcolor[RGB]{254, 248, 198}\textbf{98.14}}  & 97.97                      & 94.10                       & 97.24     & 97.43                               \\ \hline\hline 
\multicolumn{16}{c}{Houston}                                                                                                                                                                                                                                                                                                                                                                                               \\ \hline
1                          & \multicolumn{1}{c|}{Healthy grass}               & \multicolumn{1}{c|}{12}                         & 96.48  & 95.83           & 93.46           & 74.04           & \cellcolor[RGB]{254, 248, 198}\textbf{98.36}  & 93.53           & 82.69           & 85.71           & \multicolumn{1}{c|}{97.07}           & 98.53                      & 99.02                       & 98.29     & 97.31                               \\
2                          & \multicolumn{1}{c|}{Stressed grass}              & \multicolumn{1}{c|}{12}                         & 97.88  & 91.61           & 91.69           & 74.10           & \cellcolor[RGB]{254, 248, 198}\textbf{99.42}  & 98.67           & 87.37           & 93.65           & \multicolumn{1}{c|}{98.78}           & 99.43                      & 99.35                       & 87.24     & 99.27                               \\
3                          & \multicolumn{1}{c|}{Synthetic grass}             & \multicolumn{1}{c|}{6}                          & 90.62  & 97.21           & 96.92           & 67.78           & \cellcolor[RGB]{254, 248, 198}\textbf{99.85}  & 99.27           & 99.41           & 99.71           & \multicolumn{1}{c|}{97.52}           & 97.37                      & 93.87                       & 97.96     & 98.25                               \\
4                          & \multicolumn{1}{c|}{Tree}                        & \multicolumn{1}{c|}{12}                         & 90.24  & 90.08           & 94.83           & 60.16           & 88.44           & 96.47           & 90.24           & 94.02           & \multicolumn{1}{c|}{\cellcolor[RGB]{254, 248, 198}\textbf{97.30}}  & 95.98                      & 98.52                       & 94.18     & 95.49                               \\
5                          & \multicolumn{1}{c|}{Soil}                        & \multicolumn{1}{c|}{12}                         & 88.66  & 99.09           & \cellcolor[RGB]{254, 248, 198}\textbf{100.00} & 66.66           & \cellcolor[RGB]{254, 248, 198}\textbf{100.00} & 99.91           & 99.42           & 98.19           & \multicolumn{1}{c|}{98.28}           & 99.92                      & 99.01                       & 97.04     & 99.18                               \\
6                          & \multicolumn{1}{c|}{Water}                       & \multicolumn{1}{c|}{3}                          & 83.69  & 13.79           & 74.60           & 6.58            & \cellcolor[RGB]{254, 248, 198}\textbf{87.46}  & 33.54           & 71.47           & 74.61           & \multicolumn{1}{c|}{53.29}           & 46.39                      & 25.71                       & 66.46     & 56.11                               \\
7                          & \multicolumn{1}{c|}{Residental}                  & \multicolumn{1}{c|}{12}                         & 75.92  & \cellcolor[RGB]{254, 248, 198}\textbf{93.31}  & 86.63           & 42.91           & 86.47           & 81.99           & 89.04           & 93.16           & \multicolumn{1}{c|}{91.00}           & 92.60                      & 82.72                       & 86.01     & 89.47                               \\
8                          & \multicolumn{1}{c|}{Commerical}                  & \multicolumn{1}{c|}{12}                         & 58.68  & \cellcolor[RGB]{254, 248, 198}\textbf{59.75}  & 56.80           & 52.62           & 52.86           & 42.37           & 53.68           & 54.84           & \multicolumn{1}{c|}{58.11}           & 61.64                      & 59.92                       & 48.85     & 62.62                               \\
9                          & \multicolumn{1}{c|}{Raod}                        & \multicolumn{1}{c|}{12}                         & 67.86  & \cellcolor[RGB]{254, 248, 198}\textbf{83.60}  & 75.53           & 40.94           & 72.10           & 79.80           & 76.75           & 81.89           & \multicolumn{1}{c|}{78.18}           & 79.80                      & 77.36                       & 83.22     & 78.01                               \\
10                         & \multicolumn{1}{c|}{Highway}                     & \multicolumn{1}{c|}{12}                         & 63.50  & 54.03           & 75.22           & 37.48           & 76.39  & 84.12           & 66.74           & 73.57           & \multicolumn{1}{c|}{\cellcolor[RGB]{254, 248, 198}\textbf{78.80}}           & 77.14                      & 69.74                       & 85.37     & 75.15                               \\
11                         & \multicolumn{1}{c|}{Railway}                     & \multicolumn{1}{c|}{12}                         & 78.77  & 81.99           & 84.88  & 20.64           & 63.58           & 80.11           & 79.43           & 40.46           & \multicolumn{1}{c|}{\cellcolor[RGB]{254, 248, 198}\textbf{90.83}}           & 91.91                      & 87.28                       & 85.47     & 90.09                               \\
12                         & \multicolumn{1}{c|}{Parking lot 1}               & \multicolumn{1}{c|}{12}                         & 60.13  & 44.74           & 40.28           & 45.74           & 59.03           & 48.55           & 52.77           & \cellcolor[RGB]{254, 248, 198}\textbf{65.01}  & \multicolumn{1}{c|}{44.09}           & 43.59                      & 43.59                       & 69.40     & 44.83                               \\
13                         & \multicolumn{1}{c|}{Parking lot 2}               & \multicolumn{1}{c|}{4}                          & 13.28  & 0.00            & 83.66           & 11.54           & 38.12           & 61.60           & \cellcolor[RGB]{254, 248, 198}\textbf{88.88}  & 75.38           & \multicolumn{1}{c|}{20.17}           & 13.88                      & 19.74                       & 13.88     & 14.32                               \\
14                         & \multicolumn{1}{c|}{Tennis court}                & \multicolumn{1}{c|}{4}                          & 79.04  & 1.00            & 99.52           & 34.28           & 97.61           & \cellcolor[RGB]{254, 248, 198}\textbf{100.00} & \cellcolor[RGB]{254, 248, 198}\textbf{100.00} & 74.52           & \multicolumn{1}{c|}{99.76}           & 96.90                      & 99.05                       & 70.24     & 98.81                               \\
15                         & \multicolumn{1}{c|}{Running track}               & \multicolumn{1}{c|}{6}                          & 97.05  & 98.91           & \cellcolor[RGB]{254, 248, 198}\textbf{100.00} & 55.88           & 99.07           & \cellcolor[RGB]{254, 248, 198}\textbf{100.00} & 98.14           & 99.69           & \multicolumn{1}{c|}{99.69}           & 97.99                      & 98.77                       & 97.84     & 98.30                               \\ \hline
                           & \multicolumn{2}{c|}{OA}                                                                            & 77.44  & 77.86           & 82.21           & 49.81           & 80.92           & 81.00           & 80.60           & 79.82           & \multicolumn{1}{c|}{\cellcolor[RGB]{254, 248, 198}\textbf{82.53}}  & 82.00                      & 80.37                       & 81.90     & 82.11                               \\
                           & \multicolumn{2}{c|}{AA}                                                                            & 76.12  & 67.00           & 83.60           & 46.09           & 81.27           & 80.00           & \cellcolor[RGB]{254, 248, 198}\textbf{82.40}           & 80.29           & \multicolumn{1}{c|}{80.19}           & 79.54                      & 76.90                       & 78.76     & 79.68                               \\
                           & \multicolumn{2}{c|}{Kappa(\%)}                                                                     & 75.57  & 76.20           & 80.00           & 45.61           & 79.35           & 77.81           & 79.15           & 78.16           & \multicolumn{1}{c|}{\cellcolor[RGB]{254, 248, 198}\textbf{81.08}}  & 81.26                      & 78.74                       & 80.39     & 80.63                               \\ \hline \hline 
\multicolumn{16}{c}{Fanglu Tea Farm}                                                                                                                                                                                                                                                                                                                                                                                       \\ \hline
1                          & \multicolumn{1}{c|}{Massion pine}                & \multicolumn{1}{c|}{58}                         & 92.74  & 99.42           & 99.70           & 99.40           & 98.84           & 99.98           & \cellcolor[RGB]{254, 248, 198}\textbf{100.00} & 99.74           & \multicolumn{1}{c|}{99.05}           & 99.14                      & 98.65                       & 98.98     & 99.09                               \\
2                          & \multicolumn{1}{c|}{Bamboo forest}               & \multicolumn{1}{c|}{23}                         & 17.91  & 93.17           & 74.51           & 55.80           & 80.54           & 69.05           & 71.56           & \cellcolor[RGB]{254, 248, 198}\textbf{92.25}  & \multicolumn{1}{c|}{78.52}           & 77.51                      & 72.80                       & 77.20     & 79.53                               \\
3                          & \multicolumn{1}{c|}{Tea plant}                   & \multicolumn{1}{c|}{284}                        & 98.59  & 98.18           & 99.61           & 98.80           & 99.00           & \cellcolor[RGB]{254, 248, 198}\textbf{99.86}  & 99.63           & 99.35           & \multicolumn{1}{c|}{99.27}           & 99.44                      & 98.61                       & 99.61     & 99.61                               \\
4                          & \multicolumn{1}{c|}{Reed}                        & \multicolumn{1}{c|}{3}                          & 76.44  & 97.59           & \cellcolor[RGB]{254, 248, 198}\textbf{100.00} & 30.28           & 97.59           & 99.52           & \textbf{100.00} & 2.38            & \multicolumn{1}{c|}{\cellcolor[RGB]{254, 248, 198}\textbf{100.00}} & 100.00                     & 96.17                       & 100.00    & 99.52                               \\
5                          & \multicolumn{1}{c|}{Rice paddy}                  & \multicolumn{1}{c|}{68}                         & 97.94  & 99.98           & \cellcolor[RGB]{254, 248, 198}\textbf{100.00} & 99.77           & 99.55           & 99.97           & 99.98           & 99.94           & \multicolumn{1}{c|}{99.85}           & 99.94                      & 99.96                       & 99.88     & 99.79                               \\
6                          & \multicolumn{1}{c|}{Sweet potato}                & \multicolumn{1}{c|}{8}                          & 89.51  & 87.26           & 97.37           & 96.25           & 96.62           & 62.67           & 79.02           & 91.26           & \multicolumn{1}{c|}{\cellcolor[RGB]{254, 248, 198}\textbf{98.88}}  & 98.00                      & 88.76                       & 95.00     & 99.00                               \\
7                          & \multicolumn{1}{c|}{Caraway}                     & \multicolumn{1}{c|}{4}                          & 86.22  & 57.24           & \cellcolor[RGB]{254, 248, 198}\textbf{100.00} & 49.88           & 96.43           & 95.01           & 98.09           & 96.44           & \multicolumn{1}{c|}{99.29}           & 98.34                      & 97.39                       & 99.52     & 98.57                               \\
8                          & \multicolumn{1}{c|}{Weed}                        & \multicolumn{1}{c|}{19}                         & 45.80  & 88.59           & 80.85           & 62.42           & 92.92           & 90.24           & 95.50           & 82.45           & \multicolumn{1}{c|}{\cellcolor[RGB]{254, 248, 198}\textbf{96.27}}  & 93.92                      & 74.96                       & 91.28     & 88.88                               \\
9                          & \multicolumn{1}{c|}{Water}                       & \multicolumn{1}{c|}{61}                         & 99.15  & \cellcolor[RGB]{254, 248, 198}\textbf{100.00} & 98.80           & 98.90           & 98.83           & 99.98           & 98.78           & \cellcolor[RGB]{254, 248, 198}\textbf{100.00} & \multicolumn{1}{c|}{98.82}           & 98.80                      & 99.09                       & 98.80     & 98.80                               \\
10                         & \multicolumn{1}{c|}{Building road}               & \multicolumn{1}{c|}{9}                          & 97.08  & 97.87           & \cellcolor[RGB]{254, 248, 198}\textbf{100.00} & 87.79           & 98.99           & 96.97           & 99.32           & \cellcolor[RGB]{254, 248, 198}\textbf{100.00} & \multicolumn{1}{c|}{99.78}           & 100.00                     & 99.89                       & 99.32     & 99.89                               \\ \hline
                           & \multicolumn{2}{c|}{OA}                                                                            & 92.28  & 97.70           & 97.82           & 95.00           & 97.97           & 97.58           & 97.93           & 98.13           & \multicolumn{1}{c|}{\cellcolor[RGB]{254, 248, 198}\textbf{98.27}}  & 98.23                      & 96.75                       & 98.15     & 98.23                               \\
                           & \multicolumn{2}{c|}{AA}                                                                            & 80.14  & 91.93           & 95.08           & 77.93           & 95.94           & 91.32           & 94.19           & 86.38           & \multicolumn{1}{c|}{\cellcolor[RGB]{254, 248, 198}\textbf{96.97}}  & 96.50                      & 92.62                       & 95.69     & 96.26                               \\
                           & \multicolumn{2}{c|}{Kappa(\%)}                                                                     & 88.33  & 96.64           & 96.82           & 92.57           & 96.99           & 96.62           & 96.98           & 97.21           & \multicolumn{1}{c|}{\cellcolor[RGB]{254, 248, 198}\textbf{97.44}}  & 97.38                      & 95.18                       & 97.26     & 97.37                               \\ \hline
\end{tabular}
}
\label{table}
\end{table}

\subsection{Results and Performance Analysis}
\begin{figure}[!t]
    \centering
    \includegraphics[width=0.49\textwidth]{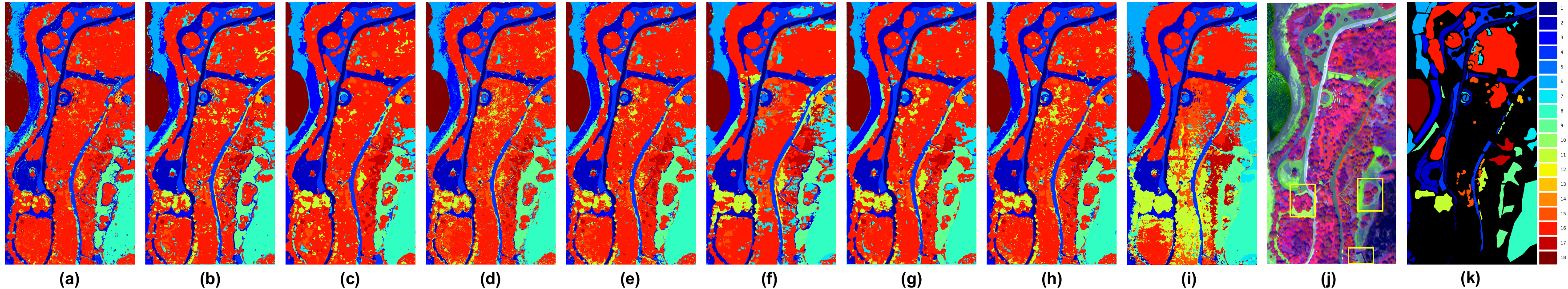}\\
    \includegraphics[width=0.49\textwidth]{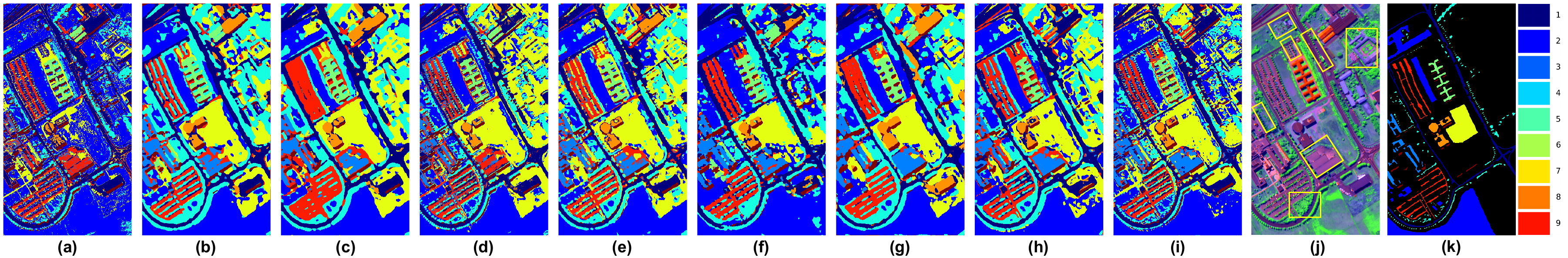}\\
    \includegraphics[width=0.49\textwidth]{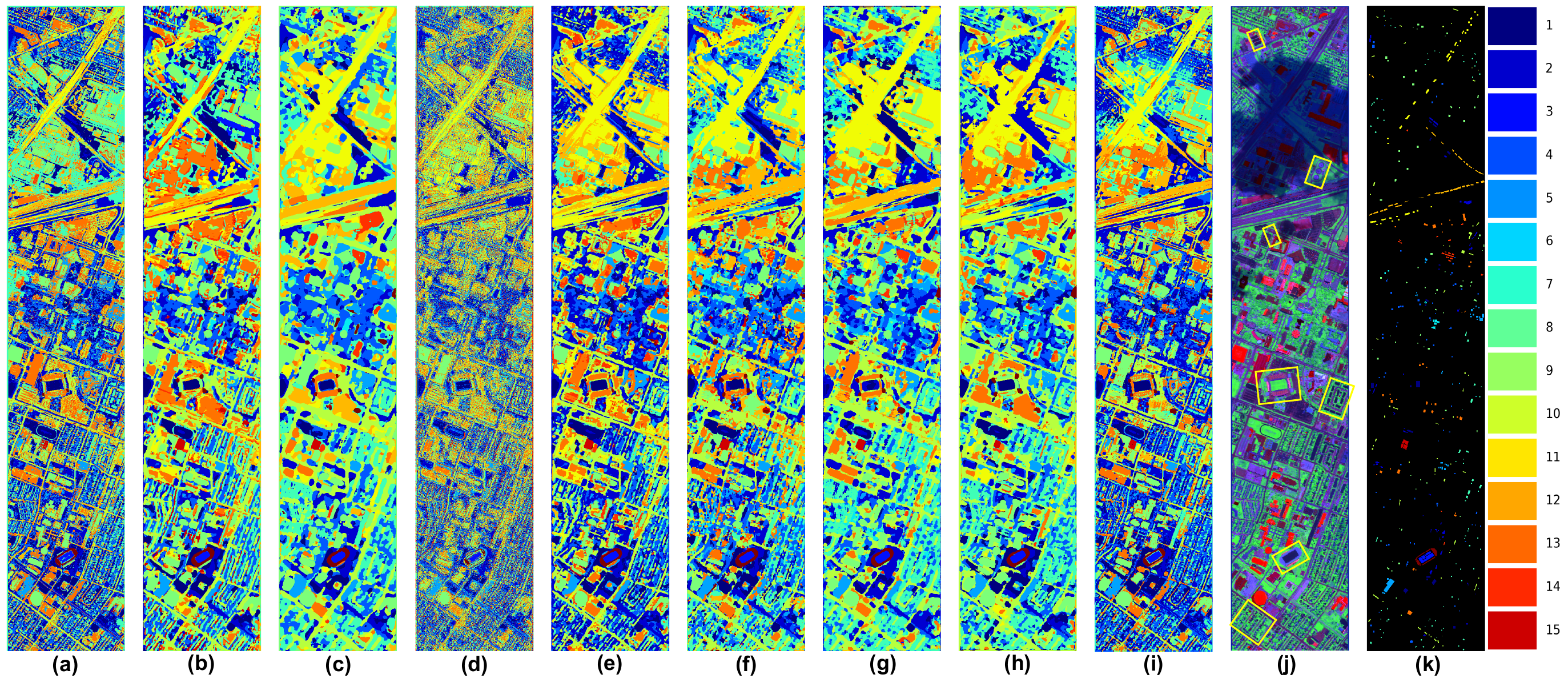}\\
    \includegraphics[width=0.49\textwidth]{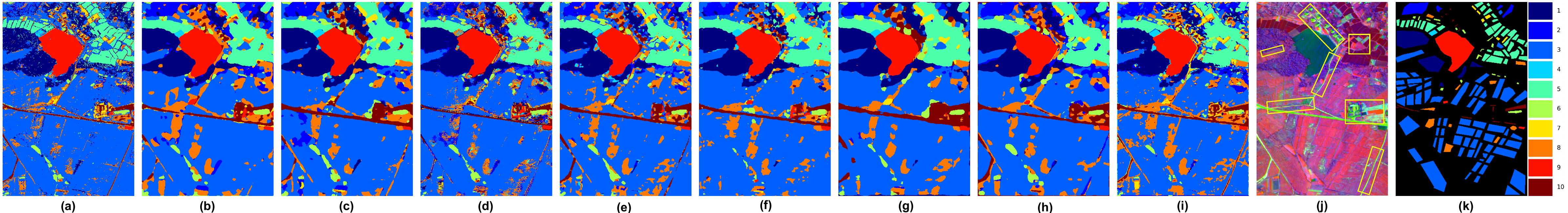}
    \caption{The classification map of (a) RF (b) SSRN (c) SS-ConvNeXt (d) DSNet (e) S\(^{2}\)VNet (f) MambaHSI (g) SFMamba (h) SDMamba (i) Ours (j) PCA image (k) Ground Truth. QUH-Tangdaowan, Pavia University, Houston, Fanglu Tea Farm, from top to bottom. Yellow boxes highlight the detailed preservation.}
    \label{maps}
\end{figure}

\autoref{table} shows the numerical results on four benchmark datasets. Overall, our proposed model achieves high accuracy in terms of OA, AA, and Kappa. Although on some datasets, for example, Houston and Fanglu Tea Farm, our model's performance is close to others, but from the visualization results, shown in \autoref{maps}, our model has outstandingly detailed boundaries. MambaHSI failed at capturing the details due to the spatial downsampling operation, although it was efficient in computation. SDMamba, a patch-based model, is limited by the sparse distributed training samples, leading to lower generalization ability, especially on the large-scale dataset, e.g., QUH-Tangdaowan dataset (1740 \(\times\) 860). Due to the large-scale and complexity of the Houston dataset, DSNet, which didn't consider the spectral variability, failed to give a satisfactory classification map, while S\(^{2}\)VNet, with endmember variability consideration, performs better than DSNet. In \autoref{maps}, our model can well capture the single trees (in Pavia University), as well as the building (in Houston) and roads (in Fanglu Tea Farm), while other models tend to give more smooth results. The ablation study (without unmixing/classification positional coding, Top-\textit{K} topken, and variability) shows that each part will enhance the performance, especially the classification positional coding, because \autoref{multi-loss} focuses more on the classification

\begin{figure}[!h]
    \centering
    \includegraphics[width=0.49\textwidth]{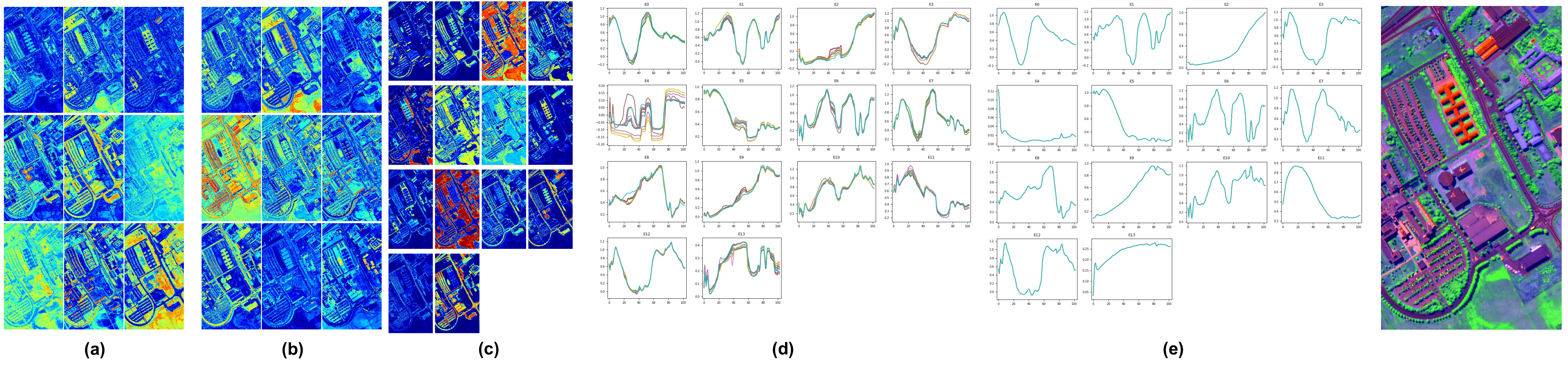}\\
    \includegraphics[width=0.49\textwidth]{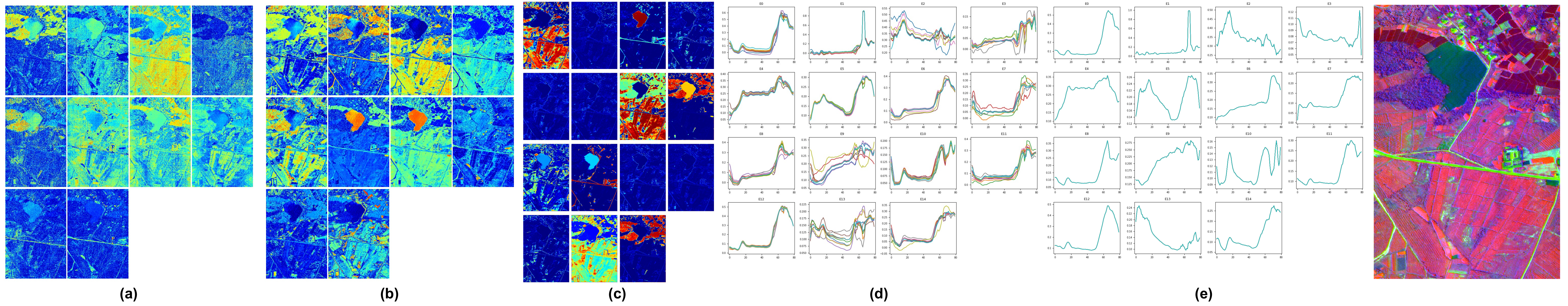}
    \caption{The estimated abundance map from (a) DSNet, (b) S\(^{2}\)VNet, (c) Ours, and the estimated (d) endmember variability, (e) Initialized endmember.}
    \label{abundance}
\end{figure}

\begin{table}[!h]
\caption{Impact of endmember number \(P\)}
\resizebox{0.49\textwidth}{!}{
\begin{tabular}{c|cccc|c|cccc}
\hline \hline
        & \multicolumn{4}{c|}{QUH-Tangdaowan} &         & \multicolumn{4}{c}{Pavia University} \\ \hline
\#Num \(P\) & 18      & 23      & 28     & 33     & \#Num \(P\) & 9       & 14      & 19      & 24     \\ \hline
OA      & 98.00   & 98.17   & 98.08  & 98.05  & OA      & 98.15   & 98.59   & 97.93   & 97.98  \\
AA      & 95.63   & 96.12   & 95.14  & 96.59  & AA      & 97.47   & 97.92   & 97.02   & 97.40  \\
Kappa   & 97.73   & 97.92   & 97.81  & 97.78  & Kappa   & 97.55   & 98.14   & 97.26   & 97.33  \\ \hline  \hline
        & \multicolumn{4}{c|}{Houston}        &         & \multicolumn{4}{c}{Fanglu Tea Farm}  \\ \hline
\#Num \(P\) & 15      & 20      & 25     & 30     & \#Num \(P\) & 10      & 15      & 20      & 25     \\ \hline
OA      & 81.65   & 81.71   & 82.53  & 81.73  & OA      & 98.12   & 98.24   & 97.76   & 98.14  \\
AA      & 79.35   & 79.06   & 80.19  & 80.35  & AA      & 95.53   & 96.16   & 94.79   & 95.93  \\
Kappa   & 80.13   & 80.19   & 81.09  & 80.23  & Kappa   & 97.21   & 97.46   & 96.67   & 97.23  \\ \hline
\end{tabular}
}
\label{endmember_P}
\end{table}


In \autoref{abundance}, it can be clearly observed that the unmixing subnet of ours can obtain more distinct abundance estimation, which provides rich spatial information and class-wise differences. This distinct spatial feature will enhance the token selection in the Mamba model, not only considering the quality (high abundance value), but also the diversity (high abundance values are distributed across the image). For example, our model gives better abundance estimation of meadows and trees (line 1 column 3), as well as the building shadows, even the shadow caused by illumination under the trees (line 1 column 1). Additionally, the road abundance in Fanglu Tea Farm is also well estimated. Additionally, the global endmember (\autoref{abundance} (e)) with adaptive variability weights \(\mathbf{w}\) can better model localized endmember variants (\autoref{abundance} (d)).

Furthermore, we evaluate the impact of the number of endmembers \(P\) on the classification performance. The results in \autoref{endmember_P} show that when \(P\) equals the number of classes, failing to achieve the best results. A higher number of endmembers means considering the spectral variability, leading to a diverse token subset for Mamba model processing, which can also further reduce the computational cost.

\section{Conclusion}  \label{conclusion}

In this letter, a new Mamba model is designed with unmixing assistance for high-quality token selection. The abundance map guides the classification branch to dynamically and adaptively choose the physically meaningful tokens. The estimated endmember reflects the spectral variability, as shown in \autoref{abundance}. Hence, the classification map has a clearer boundary, which is important for real-world land cover and land use applications. In the feature, under this framework, we will design a model that can dynamically decide the number of endmembers for better novel category discovery, and a better way to explore the spectral variability, and a way that can do so without the endmember initialization. Furthermore, we will extend this method into a domain adaptation model, as well as the foundation model.

\bibliographystyle{IEEEtran}
\bibliography{IEEEabrv,ref}

%






\end{document}